\documentclass{article}

\usepackage{microtype}
\usepackage{graphicx}
\usepackage{subfigure}
\usepackage{booktabs} 

\usepackage{amsmath}
\usepackage{amsfonts}
\usepackage{hyperref}
\usepackage{dblfloatfix}
\usepackage{enumitem}

\usepackage{algpseudocode}

\usepackage[accepted]{mlsys2024}

\mlsystitlerunning{Striped Attention}

\begin{document}
\twocolumn[
\mlsystitle{Striped Attention: \\ Faster Ring Attention for Causal Transformers}

\mlsyssetsymbol{equal}{*}

\begin{mlsysauthorlist}
\mlsysauthor{William Brandon}{mit}
\mlsysauthor{Aniruddha Nrusimha}{mit}
\mlsysauthor{Kevin Qian}{mit}
\mlsysauthor{Zachary Ankner}{mit,mosaic}
\mlsysauthor{Tian Jin}{mit}
\mlsysauthor{Zhiye Song}{miteecs}
\mlsysauthor{Jonathan Ragan-Kelley}{mit}
\end{mlsysauthorlist}

\mlsysaffiliation{mit}{MIT CSAIL, Cambridge, MA, USA}
\mlsysaffiliation{miteecs}{MIT EECS, Cambridge, MA, USA}
\mlsysaffiliation{mosaic}{MosaicML, San Francisco, CA, USA}

\mlsyscorrespondingauthor{William Brandon}{wbrandon@csail.mit.edu}

\mlsyskeywords{Machine Learning, MLSys}

\vskip 0.3in

\begin{abstract}
To help address the growing demand for ever-longer sequence lengths in transformer models, Liu et al. recently proposed Ring Attention (\citeyear{liu2023ring}), an exact attention algorithm capable of overcoming per-device memory bottlenecks by distributing self-attention across multiple devices.
In this paper, we study the performance characteristics of Ring Attention in the important special case of \emph{causal} transformer models, and identify a key workload imbalance due to triangular structure of causal attention computations. 
We propose a simple extension to Ring Attention, which we call \emph{Striped Attention} to fix this imbalance.
Instead of devices having contiguous subsequences, each device has a subset of tokens distributed uniformly throughout the sequence, which we demonstrate leads to more even workloads.
In experiments running Striped Attention on A100 GPUs and TPUv4s, we are able to achieve up to $1.45\times$ end-to-end throughput improvements over the original Ring Attention algorithm on causal transformer training at a sequence length of 256k. Furthermore, on 16 TPUv4 chips, we were able to achieve $1.65\times$ speedups at sequence lengths of 786k. We release the code for our experiments as open source.
\end{abstract}
]
\printAffiliationsAndNotice{}  

\section{Introduction}
\begin{figure*}[t]
    \centering
    \includegraphics[scale=0.8]{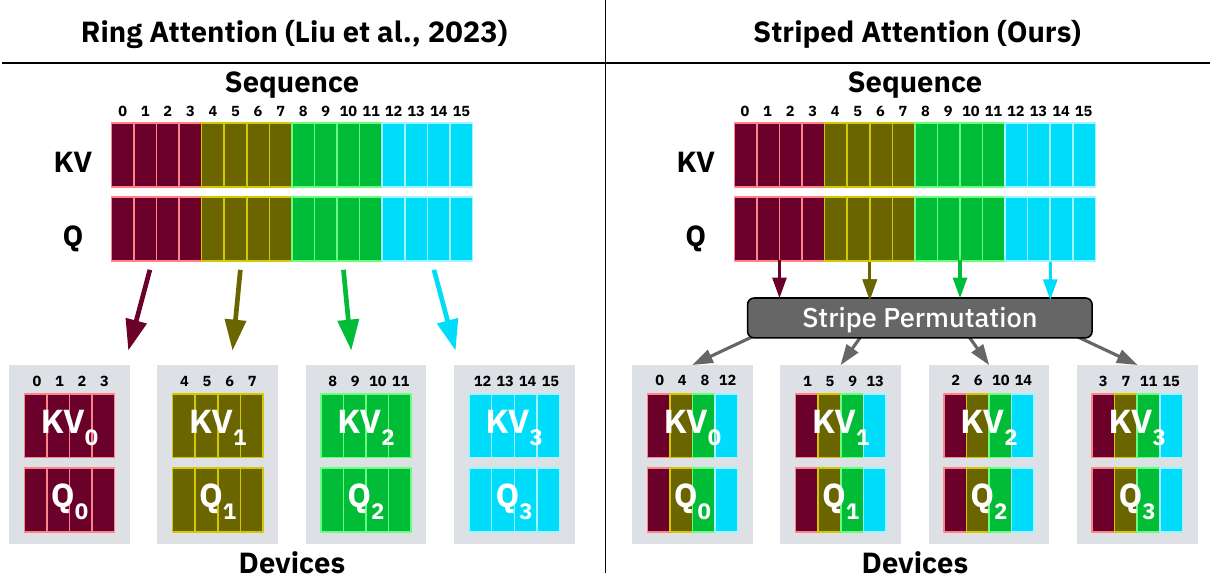}
    \caption{Initial partitioning of the $Q$, $K$, and $V$ sequences into blocks for both Ring Attention and Striped Attention. Because they travel together in the Ring Attention algorithm, the $K$ and $V$ sequences are depicted as a single sequence. Note that for both Ring Attention and Striped Attention, the tokens in the input sequence are partitioned among devices before running the first layer of the model, and remain partitioned in the same layout throughout the forward and backward passes. As a result, $Q, K, V$ are automatically partitioned in the desired layout at the beginning of each attention layer when using both Ring Attention and Striped Attention, with no extra per-layer communication required to prepare them in this state.}
    \label{fig:ring_vs_striped_partitioning}
\end{figure*}

In pursuit of the goal of training transformer models \cite{transformer} with extremely long maximum sequence lengths, Liu et al. recently proposed \emph{Ring Attention} (\citeyear{liu2023ring}), an efficient algorithm for distributing self-attention computations across multiple devices connected in a ring topology. By sharding the queries, keys, and values in each self-attention layer across the memory of multiple accelerators, Ring Attention makes it possible to train transformers on sequences which are device-count times larger than would fit on a single accelerator. Moreover, by scheduling the distributed self-attention computation in such a way that cross-device communication can be overlapped with on-device computation, Ring Attention promises high throughput and low overhead relative to implementations of self-attention which run on a single device. To the best of our knowledge, Ring Attention represents the current state of the art in efficient algorithms for exact long-context self-attention.

The purpose of this paper is to demonstrate a simple trick by which the throughput of Ring Attention can be improved even further in the particular case of \emph{causal} self-attention, 
the type of attention used in generative language models such as GPT \cite{radford2018improving,radford2019language,gpt3} and Llama \cite{touvron2023llama,touvron2023llama2}.
It is well known that causal self-attention can be computed more cheaply than general bidirectional self-attention; in causal self-attention, each query only interacts with keys which appear \emph{earlier} than it in the sequence, reducing the total number of operations required by roughly a factor of $2\times$. In the single-device setting, optimized kernels such as FlashAttention \cite{flash_attention,dao2023flashattention} already routinely exploit this fact about causal attention to deliver significantly increased throughput relative to the na\"ive strategy of indiscriminately computing all pairwise query/key interactions and then separately applying a mask to preserve only causal interactions.

Unlike existing single-device attention kernels, we observe that Ring Attention \emph{cannot} make effective use of the structure of causal attention to improve its throughput on a per device basis.
The reason for this limitation is a \textbf{workload imbalance}: on all but the first iteration of the Ring Attention algorithm, 
the workload of some devices is entirely necessary (unmasked), while the workload of others is entirely unnecessary (masked) for the final output.
The latency of RingAttention is determined by the \emph{maximum} latency of any participating device per iteration. As a result, regardless of per device optimizations, the latency per iteration would be the same as the time taken to compute a fully unmasked workload. As a result, RingAttention will run as fast as a workload with no attention masking, despite in principle needing to compute only half the operations. To successfully exploit the structure of causal self-attention, we need to change how Ring Attention partitions work among devices.

We propose \emph{Striped Attention}, a variant of Ring Attention which \textbf{permutes the input sequence} in a way which \textbf{almost entirely eliminates the workload imbalance} present in the original Ring Attention algorithm. In particular, we permute the sequence so that every device always operates on a discontiguous subset of tokens distributed approximately \emph{uniformly} throughout the original sequence. This ensures that approximately half of the query/key interactions on each device will be inhibited by the causal mask.
Like Ring Attention, Striped Attention is an \emph{exact} attention algorithm; we exploit the permutation equivariance of the core attention computation to ensure that our decision to internally permute the input sequence does not affect the final output of the model. Unlike Ring Attention, Striped Attention is able to take advantage of the structure of causal attention to save time on every iteration.

Using an implementation of Striped Attention built as an extension to Liu et al.'s Ring Attention codebase in JAX \cite{jax2018github}, we empirically observe end-to-end speedups of up to $1.45\times$ when training billion-scale causal language models on sequences with hundreds of thousands of tokens using a server with 8 A100 80GB GPUs. We see evern greater speedups on TPUs, achieving $1.65\times$ speedups on problems with over half a million tokens on 16 TPUv4 chips.
We release the code for our experiments as open source.\footnote{\href{https://github.com/exists-forall/striped_attention/}{https://github.com/exists-forall/striped\_attention/}} We hope that Striped Attention can serve as a foundational infrastructural technique to enable researchers to explore new applications in the emerging domain of extremely long-context causal transformer models.

In summary, the contributions of this paper are as follows:
\begin{itemize}[noitemsep, topsep=0pt]
    \item We identify a \textbf{workload imbalance} in Ring Attention which prevents it from taking advantage of the structure of causal attention to save computational work.
    \item We propose \textbf{Striped Attention}, a modification to Ring Attention which resolves this workload imbalance to improve throughput on causal sequence modeling tasks.
    \item We experimentally measure the throughput of causal transformer training with Striped Attention, and find that our implementation enables \textbf{up to $\mathbf{1.65\times}$ end-to-end speedups} over the original Ring Attention algorithm when training with long sequence lengths.
\end{itemize}

\section{Background}

\subsection{Causal Self-Attention}

The central goal of this paper is to define an efficient distributed implementation of the core \emph{causal self-attention} operation in the transformer architecture \cite{transformer}. Causal self-attention takes as input matrices $Q, K, V \in \mathbb{R}^{n_\text{seq} \times d_\text{head}}$, where $n_\text{seq}$ is the input sequence length and $d_\text{head}$ is a hyperparameter, and (eliding irrelevant scaling factors) computes
\begin{align*}
    \operatorname{CausalAttn}(Q, K, V) = \operatorname{Softmax}\left( Q K^\top + C \right) V
\end{align*}
where the $\operatorname{Softmax}$ is computed row-rise, and $C$ is a \emph{causal mask} matrix all of whose elements above the diagonal are $-\infty$, with zeros elsewhere:
\begin{align*}
    C_{i,j} = \begin{cases}
        -\infty & \text{if $i < j$} \\
        \phantom{-}0 & \text{if $i \ge j$}
    \end{cases}
\end{align*}
The effect of adding $-\infty$ to an entry of the matrix $Q K^\top$ is to remove it from consideration in the $\operatorname{Softmax}$, and to ensure that the output of the $\operatorname{Softmax}$ for that entry will be zero.
The causal mask can therefore be interpreted as ensuring that the output row of attention at a given position in the sequence will be affected only by rows of the $K$ and $V$ matrices occurring at equal or earlier positions in the sequence.

Inspecting the structure of the causal self-attention operation, we find two opportunities for saving work in an optimized implementation:
\begin{enumerate}[label=(\arabic*), noitemsep, topsep=0pt]
    \item We can avoid computing approximately half the elements of the product $QK^\top$: namely, those above the diagonal, which will all be set to $-\infty$ by $C$.
    \item We can avoid multiplying $V$ by approximately half the entries of $\operatorname{Softmax}(QK^\top + C)$: namely, those above the diagonal, which are all guaranteed to be zero.
\end{enumerate}
These optimizations are well-known and are utilized in fused implementations of attention like FlashAttention \cite{flash_attention,dao2023flashattention} which avoid materializing the full $\operatorname{Softmax}(QK^\top + C)$ matrix in global memory. Under ideal conditions, these optimizations can deliver up to a $2\times$ reduction in the number of FLOPs required to compute causal attention; in reality, hardware limitations and fixed overheads reduce this factor.

\subsection{Ring Attention}
\begin{figure*}[t]
    \centering
    \includegraphics[scale=0.6]{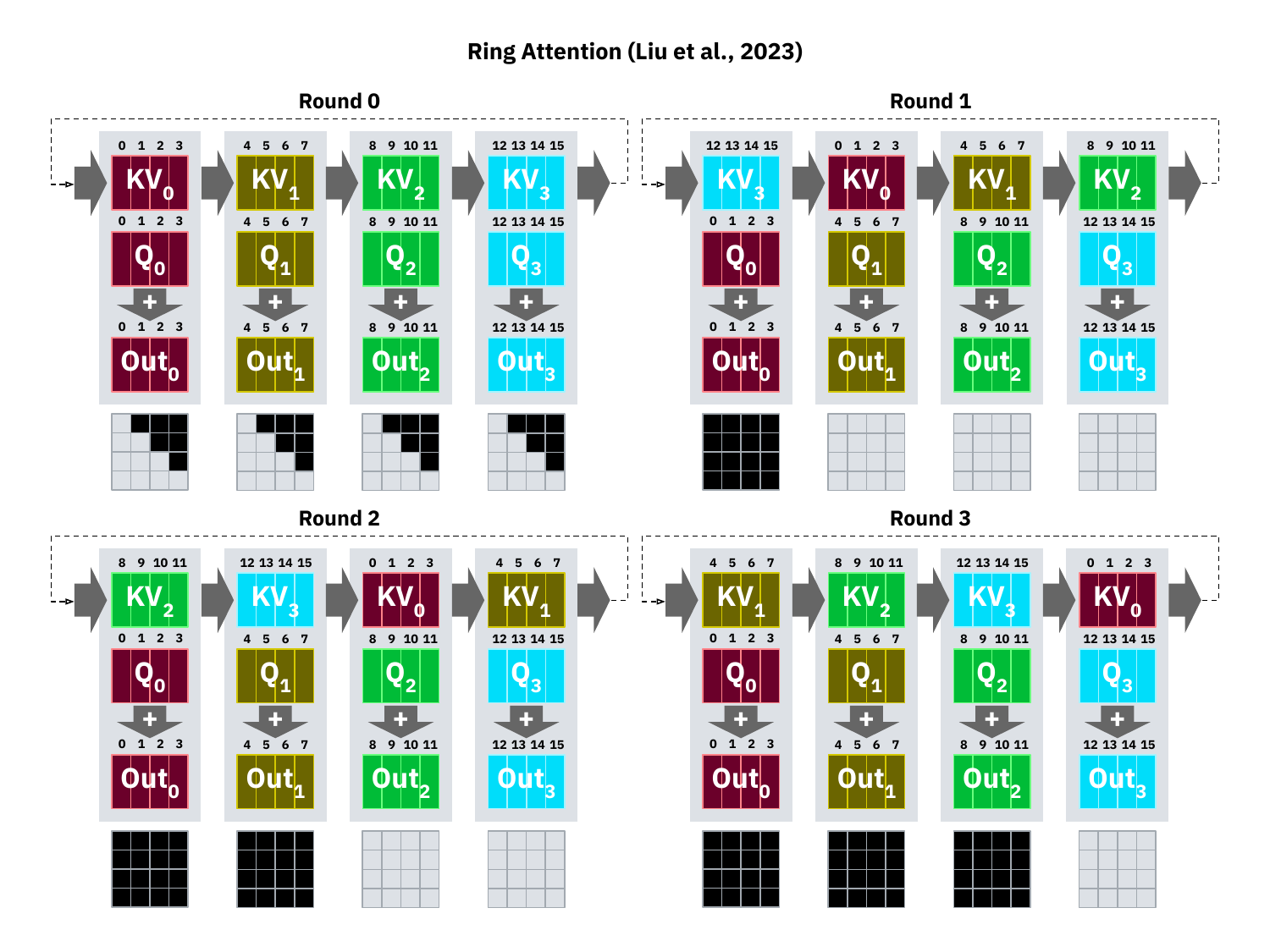}
    \caption{Behavior of Ring Attention as applied to a small causal self-attention problem with $n_\text{seq} = 16$, distributed across $N = 4$ devices. The $Q$ blocks remain stationary, while $K$ and $V$ blocks pass from neighbor to neighbor in a circular fashion on each round. The square tile shown under each device on each round indicates the causal mask for the query/key interactions computed by that device on that round; elements masked out with values of $-\infty$ are indicated in black. We can see that on all but the first round, workload imbalances prevent us from making effective use of the structure of the causal mask to reduce run time.}
    \label{fig:ring_attention_computation_all_rounds}
\end{figure*}

The recently-proposed Ring Attention algorithm \cite{liu2023ring} provides a strategy for distributing the attention computation across multiple accelerators, such as GPUs or TPUs, connected in a ring communication topology. Given $N$ available devices, Ring Attention assumes that the matrices $Q, K, V$ are initially sharded across the sequence dimension, so that they decompose into evenly-sized blocks of rows as
\begin{align*}
    Q = \begin{bmatrix}
        Q_0 \\
        \vdots \\
        Q_{N - 1}
    \end{bmatrix} \quad
    K = \begin{bmatrix}
        K_0 \\
        \vdots \\
        K_{N - 1}
    \end{bmatrix} \quad
    V = \begin{bmatrix}
        V_0 \\
        \vdots \\
        V_{N - 1}
    \end{bmatrix}
\end{align*}

\begin{algorithm*}[t]
    \caption{The Ring Attention algorithm introduced by Liu et al. (\citeyear{liu2023ring}). \\ The accumulators $\textit{Out}_0, \ldots, \textit{Out}_{N-1}$ store unnormalized partial sums of the output, as well as running softmax statistics, and the ``\textbf{normalize}'' step normalizes with respect to these statistics, as described by Rabe and Staats (\citeyear{rabe2021self}). The ``$\text{AccumulateAttentionFragment}$'' function encapsulates the logic needed to compute partial attention results on a fragment of the inputs, accumulate into the output, and update running softmax statistics. Critically, it skips computation for each tile that is masked out. The algorithm listed here serves as the backbone of both Ring Attention and Striped Attention, which differ only in how they implement the function $\text{GetMask}(j, k)$.}
    \label{alg:ringpseudocode}
\begin{algorithmic}
\Procedure{RingAttention}{$(Q_0, \ldots, Q_{N-1}), (K_0, \ldots, K_{N-1}), (V_0, \ldots, V_{N-1})$}
  \State Initialize output accumulators $\textit{Out}_0, \ldots, \textit{Out}_{N-1}$ on devices $0, \ldots, N-1$
  \For{$i = 0$ \textbf{to} $N - 1$}
    \ForAll{\textbf{devices} $j = 0$ \textbf{to} $N - 1$ \textbf{in parallel}}
      \State \textbf{let} $k \leftarrow (j - i) \bmod N$
      \State $\textit{Mask} \leftarrow \text{GetMask}(j, k)$
      \State $\textit{Out}_j \leftarrow \text{AccumulateAttentionFragment}(\textit{Out}_j, Q_j, K_k, V_k, \textit{Mask})$
      \State \textbf{send} $K_k, V_k$ \textbf{to device} $(j + 1) \bmod N$
      \State \textbf{receive} $K_{(k - 1) \bmod N}, V_{(k - 1) \bmod N}$ \textbf{from device} $(j - 1) \bmod N$
    \EndFor
  \EndFor
  \State \textbf{normalize} $(Out_0, \ldots, Out_{N-1})$
  \State \Return $(Out_0, \ldots, Out_{N-1})$
\EndProcedure
\end{algorithmic}
\end{algorithm*}

such that device $0$ initially holds the blocks $Q_0, K_0, V_0$, device $1$ initially holds $Q_1, K_1, V_1$, and so on. The left side of Figure \ref{fig:ring_vs_striped_partitioning} shows the partitioning strategy. Ring Attention then computes the output of attention by executing $N$ rounds of computation and communication, keeping the $Q$ blocks stationary on their respective devices while passing the $K$ and $V$ blocks from neighbor to neighbor in a circular fashion. On each iteration, each device computes the interactions between the $Q$ block it owns, and the $K$ and $V$ blocks it is currently holding. Each device owns a stationary block of rows of the output matrix corresponding to the same sequence positions as its $Q$ block. Devices accumulate partial outputs across iterations using the \emph{lazy softmax} strategy introduced by Rabe and Staats (\citeyear{rabe2021self}).
 
For expository purposes, pseudocode describing the high-level structure of the Ring Attention algorithm is provided in Algorithm \ref{alg:ringpseudocode}. Of particular note in this algorithm is the function $\text{GetMask}(j, k)$, which returns the tile of the attention mask governing the interaction between the query block $Q_j$ and the key/value blocks $K_k, V_k$. When using Ring Attention to compute causal self-attention, the output of $\text{GetMask}(j, k)$ is determined as shown below:

\begin{samepage}
\hrulefill
    \begin{algorithmic}
        \Procedure{GetMaskRingAttention}{$j, k$}
        \If{$j < k$}
            \State $\forall x,y \textit{ MASK}[x,y] = -\infty$
        \ElsIf{$j = k$}
            \State $\forall x,y \mid y<x \textit{ MASK}[x,y] = -\infty$
        \EndIf
        \EndProcedure
    \end{algorithmic}
\hrulefill
\end{samepage}

If we interpret the original $n_\text{seq} \times n_\text{seq}$ causal mask $C$ as an $N \times N$ matrix of blocks each of shape $\frac{n_\text{seq}}{N} \times \frac{n_\text{seq}}{N}$, this logic for $\text{GetMask}(j, k)$ is equivalent to selecting the entry at position $j, k$ in $C$.

We illustrate the behavior of the Ring Attention algorithm in the case of causal self-attention in Figure \ref{fig:ring_attention_computation_all_rounds}. We can see that on all but the first iteration, workload imbalances prevent us from usefully exploiting the structure of the causal mask to reduce run time; although some devices' interactions are entirely masked out, we cannot save any time by having these devices skip their computations because other devices' interactions are entirely unmasked.

\subsubsection{Tiling} \label{tiledesc}
For the purposes of efficient avoidance of unnecessary computation, blocks are further divided into \textit{tiles}. 
For each tile, we check if it is entirely masked out. 
If so, we skip computing the tile entirely.

For example, if we had a tile size of $512 \times 512$, and a block size of nine tiles (or $1536 \times 1536$), 
then in the case of a block with casual masking three tiles would be fully unmasked, three tiles would be partially masked, and three tiles would be fully unmasked. 
We would avoid computing the last three tiles, resulting in a $33\%$ reduction in compute. 
Note that this is less than the $50\%$ reduction in FLOPs.
However, if the block we were to compute were fully unmasked, we would need to compute all nine tiles.

\begin{figure*}[t]
    \centering
    \includegraphics[scale=0.6]{
    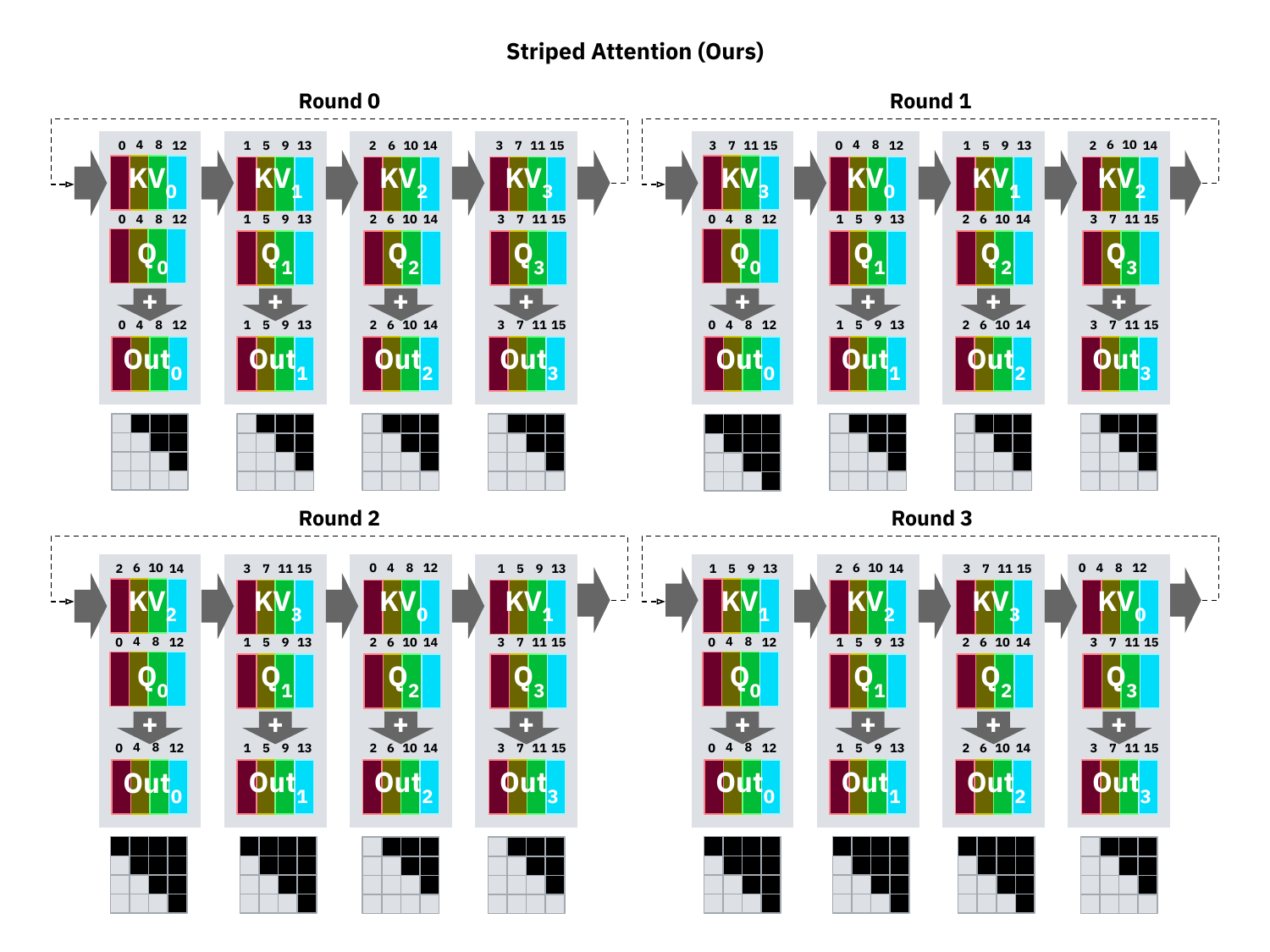}
    \caption{Behavior of Striped Attention as applied to the same casual self-attention problem as above. Unlike in Ring Attention, every block contains tokens distributed throughout every part of the original input sequence. As in Figure \ref{fig:ring_attention_computation_all_rounds}, the square matrices under each device depict the causal mask encountered by each device on each round. We note that the causal masks provide each device with a roughly equal portion of skippable work on each iteration, resolving the workload imbalance in Ring Attention.} 
    \label{fig:striped_attention_computation_all_rounds}
\end{figure*}

Striped Attention reduces the workload imbalance in Ring Attention for causal transformers by partitioning the sequence in a novel way. Rather than partitioning the tokens into contiguous blocks, we partition them into sets of evenly-spaced \emph{stripes} based on their residues modulo the device count $N$. For example, given a device count of $4$, device $0$ would own tokens $0, 4, 8, \ldots$, device $1$ would own tokens $1, 5, 9, \ldots$, and so on. Figure \ref{fig:ring_vs_striped_partitioning} illustrates this example. In practice, we can achieve this partitioning scheme by permuting the input tokens before the model's first embedding layer, and then partitioning the permuted sequence into contiguous blocks as in Ring Attention. For models which use position embedding schemes, such as RoPE \cite{rope-paper}, the position ids must be permuted as well, and in the training setting we must also permute the sequence of target token ids used to compute the loss.

After partitioning, our algorithm proceeds almost identically to Ring Attention (Figure \ref{fig:striped_attention_computation_all_rounds}). We conceptualize the devices as a ring indexed from $0$ to $N$. At the beginning of each attention layer, each device $i$ initially holds blocks $Q_i, K_i, V_i$, which inherit the striping permutation applied to the original input sequence. On each round of the Striped Attention algorithm, each device $i$ computes causal attention interactions between its assigned $Q_i$ block and the $K, V$ blocks that it received in the previous iteration. Simultaneously, it sends its current $K, V$ blocks on to the next device in the ring to hide memory latency.

\begin{figure}
    \centering
    \subfigure[Workload distribution in Ring Attention]{\includegraphics[width=\columnwidth]{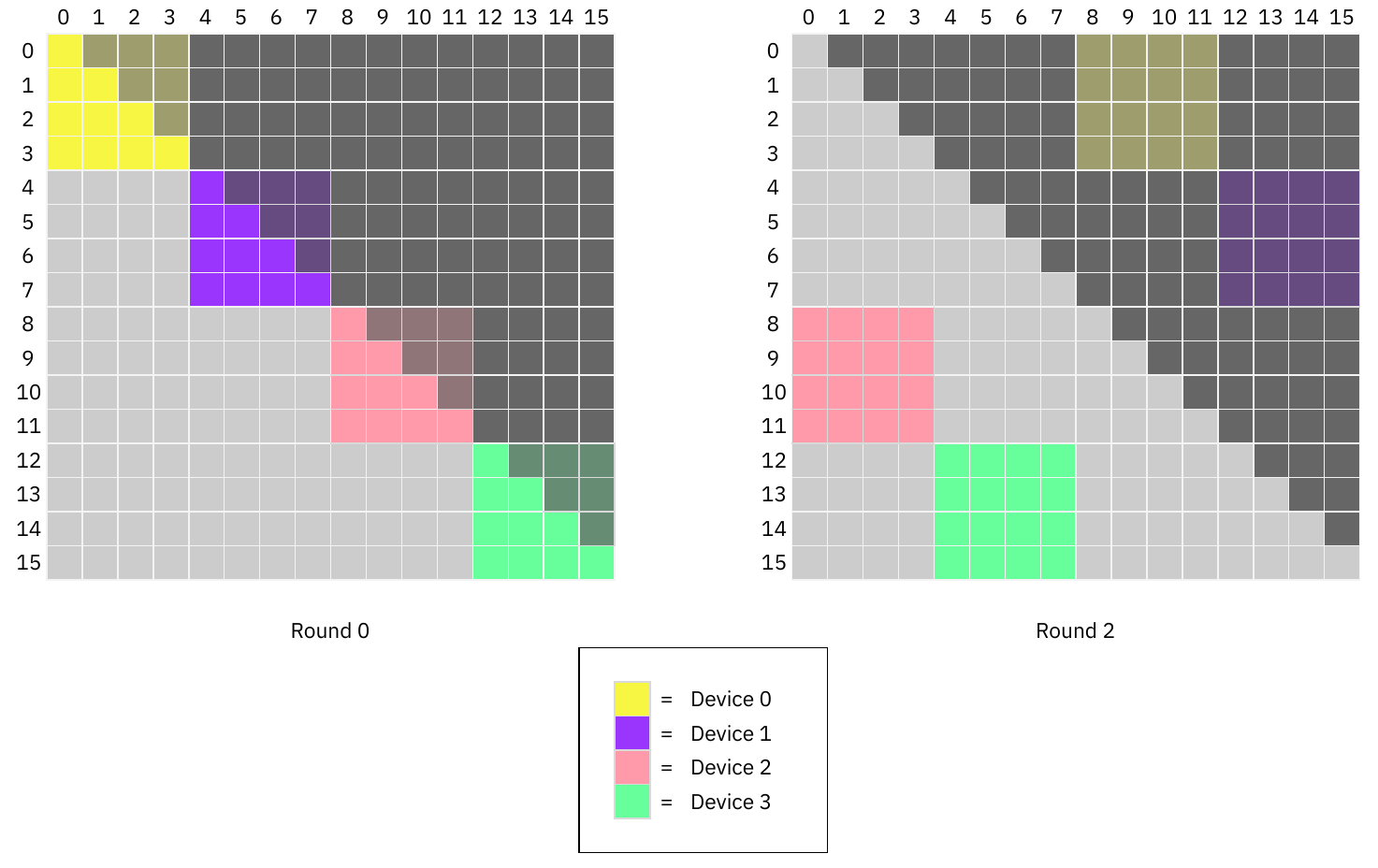}} %
    \subfigure[Workload distribution in Striped Attention]{\includegraphics[width=\columnwidth]{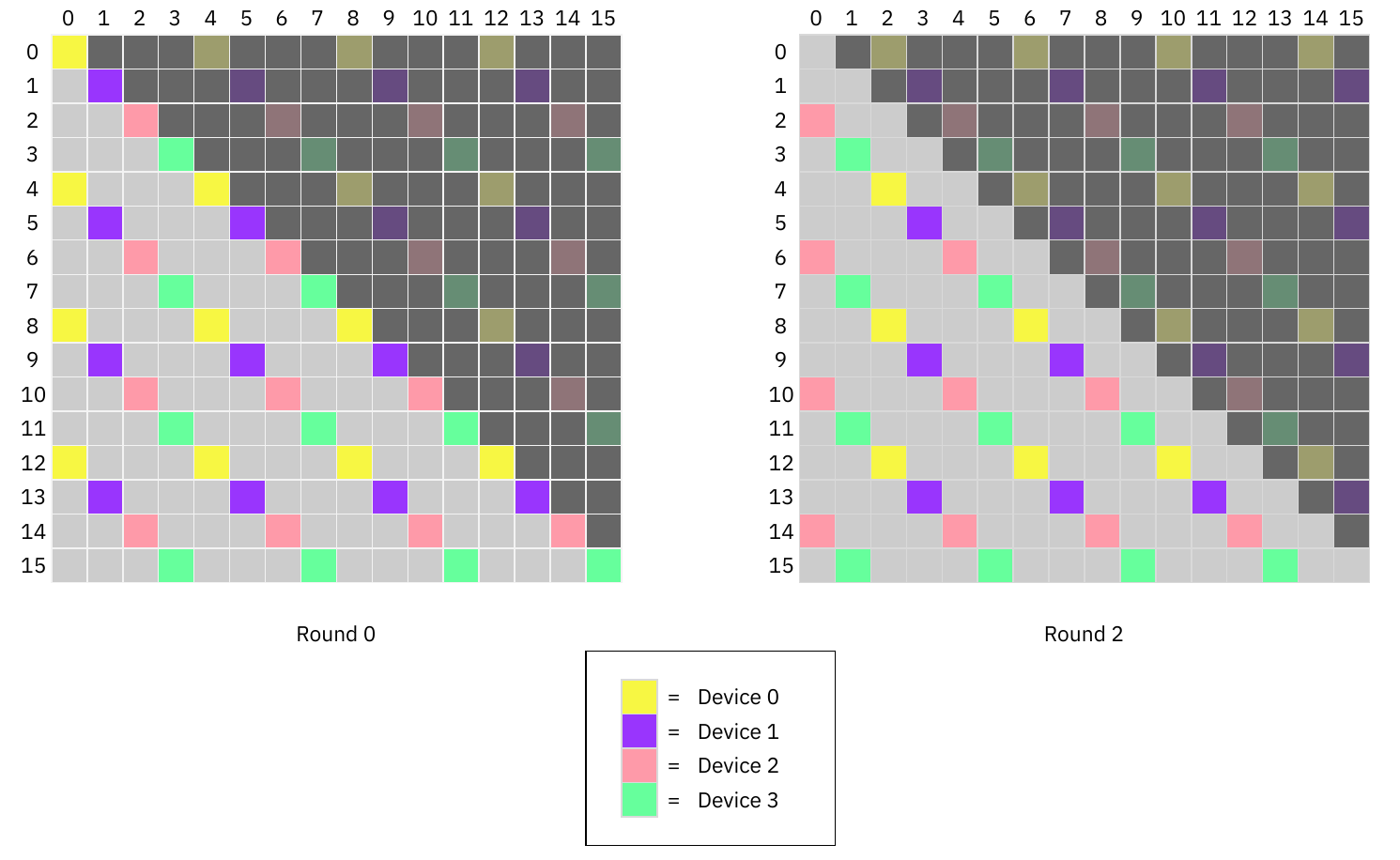}}
    \caption{Workload distribution on rounds 0 and 2 of Ring Attention and Striped Attention. The square matrices represent the set of all possible pairwise query/key interactions; row indices correspond to queries, and column indices correspond to keys. All cells above the diagonal are masked out by the causal mask and can be skipped. The colors indicate which devices are responsible for which parts of the computation. On round 2, we can see that some devices in Ring Attention are responsible for workloads which are entirely masked out, whereas Striped Attention maintains a balanced workload across all devices.}
    
    \label{fig:striped_attention_work}
\end{figure}

Crucially, the attention computation in Striped Attention is causal with respect to the order in which tokens appear in the \emph{original input sequence}, not their order in the permuted sequence. This has important implications for the structure of the mask returned by $\text{GetMask}(j, k)$ (see Algorithm \ref{alg:ringpseudocode}): in contrast to Ring Attention, Striped Attention ensures that \emph{every} device's causal mask is upper-triangular on \emph{every} iteration (see Figure \ref{fig:striped_attention_work}). The Algorithm below describes the modification to $\text{GetMask}(j, k)$ function required to adapt Ringed Attention into Striped Attention.

\begin{samepage}
\hrulefill
    \begin{algorithmic}
    \Procedure{GetMaskStripedAttention}{$j,k$}
        \State Initialize $\textit{MASK} \in \mathbb{R}^{\frac{n_\text{seq}}{N}\times \frac{n_\text{seq}}{N}}$ 
        \State $\forall x,y \textit{ MASK}[x,y] = 0$ 
        \If{$j < k$}
            \State $\forall x,y \mid y \leq x \textit{ MASK}[x,y] = -\infty$
        \ElsIf{$j \geq k$}
            \State $\forall x,y \mid y<x \textit{ MASK}[x,y] = -\infty$
        \EndIf
        \State \Return \textit{MASK}
        \EndProcedure
    \end{algorithmic}
    \hrulefill
\end{samepage}

When computing causal blockwise attention between block $Q_i$ and blocks $K_j$, $V_j$, Striped Attention yields a per-device workload of
\[ \mathrm{Work}(i, j) = \begin{cases} \frac{c(c+1)}2 & i \ge j \\ \frac{c(c-1)}2 & i < j \end{cases}\]
attention interactions, where $c$ is the per-device block size. In comparison, Ring Attention's computation schedule has at least one device on each iteration which must process a workload of size $c^2$, which acts as the limiting factor on the latency of each iteration. 

As shown in the equation above, the workloads of different devices are not exactly identical; they can differ in whether or not they include the diagonal. However, as the block size grows, the fraction of the attention matrix taken up by the diagonal shrinks linearly, and the workload imbalance becomes negligible as a fraction of the total runtime. In the limit as block size and device count go to infinity, the maximum theoretical speedup from using Striped Attention over Ring Attention approaches $2\times$.

\section{Experimental Evaluation}

\subsection{Implementation}

We implement Striped Attention as an extension to Liu et al.'s Ring Attention codebase in JAX, and compare its throughput to Ring Attention on a set of billion-parameter-scale causal transformer training benchmarks. 
We run experiments on a few configurations:

\begin{enumerate}[label=(\arabic*), noitemsep, topsep=0pt]
    \item a server with 8 NVIDIA A100 80GB GPUs connected with NVLink
    \item a TPU Pod Slice with 4 TPU v3 chips, or 8 Tensor cores, and
    \item a TPU Pod Slice with 16 TPU v4 chips, or 32 Tensor cores.
\end{enumerate}

Both our Ring Attention baseline and our implementation of Striped Attention use a coarse-grained tiling strategy to skip masked out portions of the attention computation. 
On each device in each round, the implementation decomposes the space of query/key interactions into \textit{tiles} (See \ref{tiledesc}  for an explanation).
The tile size varies between our GPU and TPU experiments. We performed a sweep of tile sizes for Ring Attention, and chose the best performing values.
For both GPUs and TPUs, we found the tile sizes which performed best for Ring Attention also performed best for Striped Attention.
Specifically, for A100 80GB GPUs we chose a tile size of $2048$ queries $\times$ $4096$ keys. 
For both v3 and v4 TPUs, the tile size was $2048$ queries $\times$ $2048$ keys.
For each tile, the algorithm computes the attention interactions for the tile if and only if it contains at least one unmasked element. 
As we remarked in Section 2, for Ring Attention this work-skipping technique only confers a performance benefit on the first round of the algorithm, whereas for Striped Attention it can confer a benefit on every round.

Following the defaults suggested in the Ring Attention codebase, we perform all computations in bfloat16 format \emph{except} the core attention computation $\operatorname{Softmax}(Q K^\top + C) V$; in the core attention computation, we promote $Q, K, V$ to float32 buffers.
On NVIDIA A100 GPUs,  the matrix multiplications for attention are performed in NVIDIA TensorFloat-32 precision.
On TPUs, these operations are performed in bfloat16 precision.

In our experiments, we explore the effect of varying three aspects of the training configuration on the relative throughput of Striped Attention to Ring Attention across devices:
\begin{enumerate}[label=(\arabic*), noitemsep, topsep=0pt]
    \item \textbf{Model Size:} We investigate 1B, 3B, and 7B model configurations which are used in the original Ring Attention implementation. The hyperparameters of these models are given in Table \ref{tab:hyperparams}.
    \item \textbf{Sequence Length:} We investigate training with total cross-device sequence lengths ranging from $8192$ (8k) to $262144$ (256k).
    \item \textbf{Distributed Mesh Dimensions:} We investigate degrees of sequence parallelism ranging from $N = 2$ to $N = 8$ devices. We also investigate varying the degree of model parallelism \cite{megatronlm} between $1$ and $4$.
\end{enumerate}

\begin{table}
    \centering
    \begin{tabular}{c|ccccc}
        Model & $n_\text{vocab}$ & $d_\text{model}$ & $d_\text{ff}$ & $n_\text{layer}$ & $n_\text{head}$ \\
        \hline
        1B & 32000 & 2048 & 5504 & 22 & 16 \\
        3B & 32000 & 3200 & 8640 & 26 & 32 \\
        7B & 32000 & 4096 & 11008 & 32 & 32
    \end{tabular}
    \caption{Architectural hyperparameters for 1B, 3B, and 7B model configurations taken from the original Ring Attention codebase.}
    \label{tab:hyperparams}
\end{table}

\subsection{Results}\label{sec:results}
\begin{figure*}
    \centering
    \includegraphics[scale=0.8]{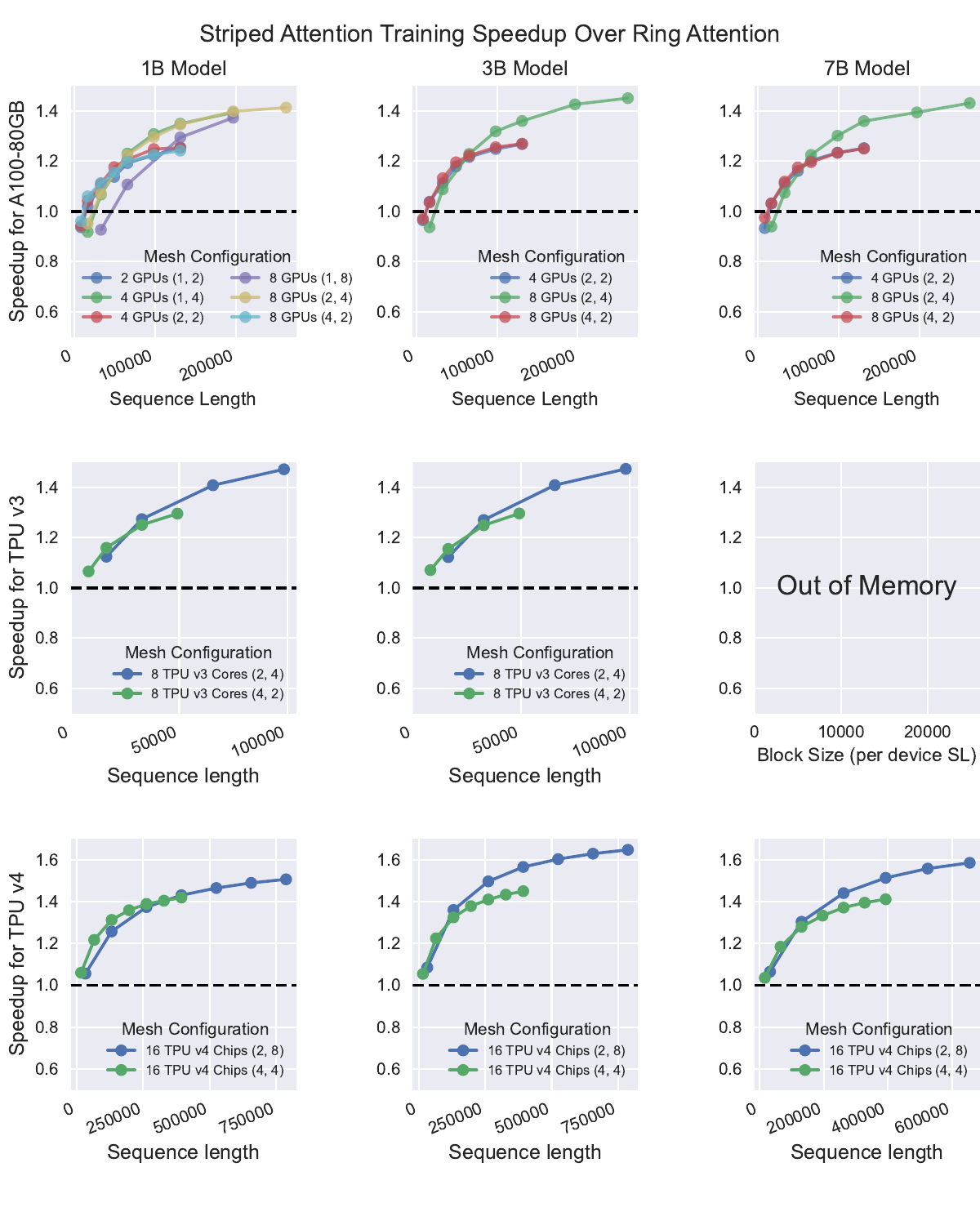}
    \caption{End-to-end training speedups of Striped Attention over Ring Attention across a range of sequence lengths and mesh configurations. The two elements of the ``Mesh Configuration'' tuple refer to the degree of model parallelism and sequence parallelism, respectively. Given a mesh configuration, increasing sequence length also increases the number of tiles per block. We generally scaled experiments until the system ran out of memory during testing.}
    \label{fig:speedup_by_seq}
\end{figure*}

\begin{figure*}
    \centering
    \includegraphics[scale=0.7]{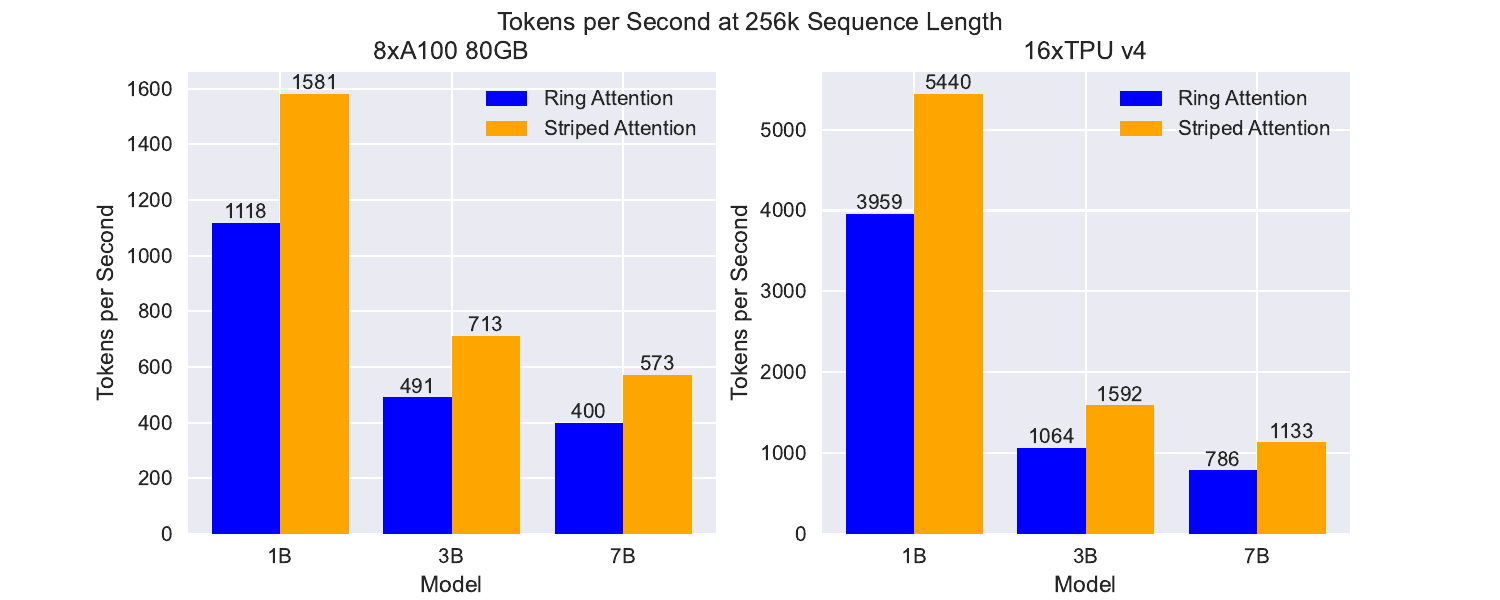}
    \caption{End-to-end training throughput from using Ring Attention and Striped Attention on our three causal language model configurations at a sequence length of 256k. All A100 experiments were done using a (2,4) mesh configuration, while all TPUv4 experiments were done with a (2,8) mesh configuration}
    \label{fig:abs_perf}
\end{figure*}

In Table \ref{tab:topresults}, we summarize the training speedups achieved by Striped Attention relative to Ring Attention for each model at a sequence length of 256k (262144). In Figure \ref{fig:speedup_by_seq}, we plot the speedups achieved over a range of sequence lengths and mesh configurations on both A100 GPUs and TPUs. Figure \ref{fig:abs_perf} plots the absolute training throughput in tokens per second achieved by each model at a sequence length of 256k.

In all configurations, the first datapoint represents a per-device block size equal to $4096 \times 4096$. 
 Since we decide whether to skip computation on the tile granularity, and the tile size for our GPU experiments is $2048 \times 4096$, in our GPU experiments Striped Attention does not save computation at the smallest block size.
 However, because the tilesize for TPUs is $2048 \times 2048$, TPUs can save 25\% of the attention FLOPs even at the smallest block size.

The most clear trend is that the benefit of our method scales with sequence length in multiple ways.
First, given a particular mesh configuration, the speedup increases with the block size. 
As the number of tiles per block increases, a greater percentage of computation is skipped, and this leads to real speedups.
Second, given a particular block size, increased sequence parallelism leads to greater speedup as long as the block has at least two tiles in the query and key dimension.
This is most clear on the A100 plot. At a given block size, all models with similar degrees of sequence parallelism have similar speedups regardless of the degree of model parallelism. However, models with high degrees of sequence parallelism have greater speedups.

\begin{table}
    \centering
    \begin{tabular}{ccr|cc}
        Model & Mesh & $n_\text{seq}$ & A100 & TPUv4\\
        \hline
        1B & $(2, 4/8)$ & 262144 & $1.41$  & $1.39$ \\
        3B & $(2, 4/8)$ & 262144 & $1.45$ & $1.50$ \\
        7B & $(2, 4/8)$ & 262144 & $1.43$ & $1.44$
    \end{tabular}
    \caption{End-to-end training speedups from using Striped Attention over Ring Attention at 256k sequence length. The two elements of the ``Mesh'' tuple refer to the degree of model parallelism and sequence parallelism, respectively. As our TPUv4 testbed had twice the devices as our A100 testbed, we doubled the degree of sequence parallelism.}
    \label{tab:topresults}
\end{table}

\section{Discussion}

In principle, Striped Attention can reduce the run time required by the core attention computation in Ring Attention by up to a factor of $2$. Considering that there are costs other than attention which contribute to run time, and that Striped Attention only speeds up the final $N - 1$ iterations of Ring Attention, for each configuration in our experiments we can quantify an effective upper bound on the speedup achievable by replacing Ring Attention with Striped Attention. We refer to this as ``Theoretical Maximum Speedup,'' or TMS. To compute TMS, we assume that all communication is perfectly overlapped with compute, and that the time taken by all non-matrix-multiply FLOPs is negligible. For our A100 GPU experiments, we assume that attention FLOPs (performed in TensorFloat-32) are twice as expensive as all other FLOPs (performed in bfloat16). For our TPU experiments (where all our matrix multiplications use bfloat16 precision) we assume all FLOPs have equal cost.

We compare our Theoretical Maximum Speedup to our empirically achieved speedup in Tables \ref{tab:gpu_tms_real_by_model}, \ref{tab:gpu_tms_real_by_mesh_sp}, and \ref{tab:tpu_tms_real_by_model}. We can see that there is a gap between the speedup achieved by our implementation of Striped Attention and the effective upper bound set by our TMS calculation. We also see that this gap narrows as sequence length increases. We attribute this gap to two main factors.

First, the fact that our implementation skips work at the granularity of \emph{tiles} means that at small per-device block sizes, we are not able to take full advantage of the causal mask to skip half the work in attention. For instance, as mentioned in Section \ref{sec:results}, when the tile size is 2048 queries $\times$ 2048 keys, and the per-device block size is $4096$, our implementation only skips computing $25\%$ of the attention matrix, even though approximately $50\%$ of the work is skippable in principle due to causal masking. As sequence length increases and the per-device block size becomes larger relative to the tile size, the overhead of tile-based work-skipping relative to ideal work-skipping becomes smaller. We look forward to future implementations of Striped Attention which employ fused kernels like FlashAttention \cite{dao2023flashattention}, which we expect will make it possible to achieve greater speedups at smaller block sizes by skipping work at a finer granularity. 

Additionally, we suspect that non-attention-computation overheads in both the Ring Attention and Striped Attention implementations we use for our experiments reduce the speedup achievable by Striped Attention. In particular, we suspect that both our Ring Attention and Striped Attention implementations may not be optimally overlapping computation and communication, and that improved overlapping may increase the observed speedup. This may contribute to explaining why our observed speedups do not appear to converge to our calculated TMS values even at large sequence length, as both computation costs and communication overheads scale with sequence length. We leave the development of more carefully optimized Ring Attention and Striped Attention implementations to future work.

\begin{table}[t]
\begin{tabular}{r|r|r|r|r}
Model& Mesh & $n_\text{seq}$ & TMS &Speedup\\
\hline
1B & $2 \times 4$ &
16384 & $ 1.46$ & $ 0.95$ \\
& &
32768 & $ 1.57$ & $ 1.07$ \\
& &
65536 & $ 1.65$ & $ 1.22$ \\
& &
98304 & $ 1.68$ & $ 1.30$ \\
& &
131072 & $ 1.70$ & $ 1.35$ \\
& &
196608 & $ 1.71$ & $ 1.40$ \\
& &
262144 & $ 1.72$ & $ 1.41$ \\
\hline
3B & $2 \times 4$ &
16384 & $ 1.38$ & $ 0.94$ \\
& &
32768 & $ 1.51$ & $ 1.09$ \\
& &
65536 & $ 1.61$ & $ 1.23$ \\
& &
98304 & $ 1.65$ & $ 1.32$ \\
& &
131072 & $ 1.67$ & $ 1.36$ \\
& &
196608 & $ 1.69$ & $ 1.43$ \\
& &
262144 & $ 1.71$ & $ 1.45$ \\
\hline
7B & $2 \times 4$ &
16384 & $ 1.34$ & $ 0.94$ \\
& &
32768 & $ 1.47$ & $ 1.07$ \\
& &
65536 & $ 1.58$ & $ 1.23$ \\
& &
98304 & $ 1.62$ & $ 1.30$ \\
& &
131072 & $ 1.65$ & $ 1.36$ \\
& &
196608 & $ 1.68$ & $ 1.39$ \\
& &
262144 & $ 1.70$ & $ 1.43$ \\
\hline
\end{tabular}
\caption{Theoretical and observed speedup for Striped Attention over Ring Attention on 8 A100 GPUs across model sizes. TMS stands for Theoretical Maximum Speedup.}
\label{tab:gpu_tms_real_by_model}
\end{table}

\begin{table}[t]
\begin{tabular}{r|r|r|r|r}
Model & Mesh & $n_\text{seq}$ & TMS & Speedup \\
\hline
1B & $2 \times 8$ &
32768 & $ 1.54$ & $ 1.06$ \\
& &
131072 & $ 1.76$ & $ 1.26$ \\
& &
262144 & $ 1.81$ & $ 1.37$ \\
& &
393216 & $ 1.83$ & $ 1.43$ \\
& &
524288 & $ 1.84$ & $ 1.47$ \\
& &
655360 & $ 1.85$ & $ 1.49$ \\
& &
786432 & $ 1.85$ & $ 1.51$ \\
\hline
3B & $2 \times 8$ &
32768 & $ 1.45$ & $ 1.09$ \\
& &
131072 & $ 1.71$ & $ 1.36$ \\
& &
262144 & $ 1.78$ & $ 1.50$ \\
& &
393216 & $ 1.81$ & $ 1.57$ \\
& &
524288 & $ 1.83$ & $ 1.60$ \\
& &
655360 & $ 1.84$ & $ 1.63$ \\
& &
786432 & $ 1.84$ & $ 1.65$ \\
\hline
7B & $2 \times 8$ &
32768 & $ 1.40$ & $ 1.07$ \\
& &
131072 & $ 1.67$ & $ 1.30$ \\
& &
262144 & $ 1.76$ & $ 1.44$ \\
& &
393216 & $ 1.80$ & $ 1.51$ \\
& &
524288 & $ 1.81$ & $ 1.56$ \\
& &
655360 & $ 1.83$ & $ 1.59$ \\
\hline
\end{tabular}
\caption{Effects of scaling up model size on the ratio of of theoretical to practical speedups on 16 TPUv4s. TMS stands for Theoretical Maximum Speedup}
\label{tab:tpu_tms_real_by_model}
\end{table}

\begin{table}
\begin{tabular}{r|r|r|r|r}
 Model & Mesh & Block Size & TMS & Speedup \\
\hline
1B & $1 \times 2$ &
4096 & $ 1.22$ & $ 0.94$ \\
& &
8192 & $ 1.31$ & $ 1.02$ \\
& &
16384 & $ 1.38$ & $ 1.10$ \\
& &
24576 & $ 1.41$ & $ 1.14$ \\
\hline
1B & $1 \times 4$ &
4096 & $ 1.46$ & $ 0.92$ \\
& &
8192 & $ 1.57$ & $ 1.07$ \\
& &
16384 & $ 1.65$ & $ 1.23$ \\
& &
24576 & $ 1.68$ & $ 1.31$ \\
\hline 
1B & $1 \times 8$ &
4096 & $ 1.67$ & $ 0.93$ \\
& &
8192 & $ 1.76$ & $ 1.11$ \\
& &
16384 & $ 1.81$ & $ 1.29$ \\
& &
24576 & $ 1.83$ & $ 1.37$ \\
\end{tabular}
\caption{Theoretical and observed speedup for Striped Attention over Ring Attention on 8 A100 GPUs across model sizes. TMS stands for Theoretical Maximum Speedup.}
\label{tab:gpu_tms_real_by_mesh_sp}
\end{table}

\section{Related Work}

\subsection{Distributed Execution in Deep Learning}
As models have grown in parameter count and computation cost, new parallelism strategies have emerged.
\citet{dean2012data} was an early example of data parallelism through a parameter server for deep learning. Model parallelism, specifically tensor parallelism, has been used at least since AlexNet \cite{alexnet}, but gained popularity for training large language models with better implementations  such as Megatron-LM \cite{megatronlm}.
These implementations achieved better scaling with tensor parallelism than prior data parallel approaches had.
As the memory cost of LLM training grew in tandem with batch sizes used during training, gradient accumulation became popular. 
This enabled efficient pipeline parallelism methods such as GPipe \cite{gpipe2018}, which was less compute efficient than prior methods but saved on memory and communication.
Fully sharded data parallelism \cite{fsdp2021}, achieved the memory savings of tensor and pipeline parallelism while sharding the parameters instead of communicating activations. 
Most relevant to our work, \cite{li2022sequence} introduced sequence parallelism, parallelism specifically targeted at long sequence length attention.
\cite{korthikanti2022reducing} proposed an efficient alternating combination of tensor and sequence parallelization, further adapting Megatron-LM to very long sequence lengths. 

\subsection{Efficient Attention Implementations}
While the feedforward layers of transformers easily achieve high utilization on modern accelerators, attention has historically had lower device efficiency. Many approaches have attempted to close this gap.
Perhaps the most common approach is the introduction of computational approximations of attention. 
While approaches such as Linformer \cite{linformer2020} and the Sparse Transformer \cite{sparsetransformer2019} attempted to use approximate the computation of attention with linear and sparse equivalents, such approaches never achieved the popularity of the original attention implementation.
\cite{rabe2021self} proposed a memory efficient algorithm for self attention that reduced the memory requirement from $O(n^2)$ to $O(1)$, as well as a practical implementation that required $O(\sqrt{n})$ memory.
FlashAttention \cite{flash_attention} provided an efficient implementation of this algorithm that leverage custom CUDA kernels and several optimizations accounting for the GPU memory hierarchy.
FlashAttention2 \cite{dao2023flashattention} further optimized the parallelization of computations across the GPU process hierarchy, better dividing work between and within thread blocks. 

\subsection{System optimizations for very long sequence lengths}
In recent months, work concurrent to this has explored other efficient distributed attention approaches for long sequence lengths.
Deepspeed Ulysses \cite{jacobs2023deepspeed} focuses on improving the communication efficiency of sequence parallel models by replacing all-gather and reduce scatter operations with smaller all-to-all operations. 
Lightseq \cite{li2023lightseq} also optimizes the communication operators for sequence parallel models. In addition, they also add an improved rematerialization strategy and enable the overlapping of communication and computation by splitting the workload into bubbles.
However, while Ring and Striped Attention keep the queries fixed while enabling the overlapping of communication and computation, the overlapping of lightseq requires moving both the queries and the keys.

\section{Conclusion}
In this work we identify and solve a workload imbalance in the recently-proposed Ring Attention algorithm for distributed long-sequence attention. 
In our experiments, we find that this solution, which we call Striped Attention, leads to speedups of up to $1.65\times$ when training causal transformers on extremely long sequences.
Furthermore, our approach is easy to implement as an extension to Ring Attention, requiring only a one-time permutation of the input sequence at the beginning of the forward pass, and an adjustment to the structure of the attention mask.

In its broader implications, Striped Attention allows for compute-efficient expansion of large language models to longer sequence lengths than have previously been explored. As our technique focuses on the case of causal attention, it is of direct relevance to training and inference for generative language models. Because Striped Attention is an \emph{exact} attention algorithm, applications can use it without making any tradeoffs in accuracy. We hope to see others build on Striped Attention to develop new applications in the emerging domain of extremely long-context generative transformer models.

\section{Acknowledgements}

First, we would like to sincerely thank Prof. Michael Carbin's lab for sponsoring the Systems for ML Discussion Group at MIT, where the idea for this paper was first conceived. We also thank Morph Labs and the Google TPU Research Cloud for providing the compute resources we used to run our experiments. We thank the MIT-IBM Watson AI Lab for providing funding which supported this research.

\bibliography{main}
\bibliographystyle{mlsys2024}

\newpage

\appendix
\section{Full results}
\subsection{A100-80GB 8 GPU Results}
\begin{tabular}{rrr|r|r}
Model & Mesh & $n_\text{seq}$ & TMS & Speedup \\
\hline
1B & $1 \times 2$ &
8192 & $ 1.22$ & $ 0.94$ \\
& &
16384 & $ 1.31$ & $ 1.02$ \\
& &
32768 & $ 1.38$ & $ 1.10$ \\
& &
49152 & $ 1.41$ & $ 1.14$ \\
& &
65536 & $ 1.43$ & $ 1.19$ \\
& &
98304 & $ 1.45$ & $ 1.22$ \\
& &
131072 & $ 1.46$ & $ 1.25$ \\
\hline
1B & $1 \times 4$ &
16384 & $ 1.46$ & $ 0.92$ \\
& &
32768 & $ 1.57$ & $ 1.07$ \\
& &
65536 & $ 1.65$ & $ 1.23$ \\
& &
98304 & $ 1.68$ & $ 1.31$ \\
& &
131072 & $ 1.70$ & $ 1.35$ \\
& &
196608 & $ 1.71$ & $ 1.39$ \\
\hline
1B & $1 \times 8$ &
32768 & $ 1.67$ & $ 0.93$ \\
& &
65536 & $ 1.76$ & $ 1.11$ \\
& &
131072 & $ 1.81$ & $ 1.29$ \\
& &
196608 & $ 1.83$ & $ 1.37$ \\
\hline
1B & $2 \times 2$ &
8192 & $ 1.22$ & $ 0.94$ \\
& &
16384 & $ 1.31$ & $ 1.04$ \\
& &
32768 & $ 1.38$ & $ 1.11$ \\
& &
49152 & $ 1.41$ & $ 1.18$ \\
& &
65536 & $ 1.43$ & $ 1.21$ \\
& &
98304 & $ 1.45$ & $ 1.25$ \\
& &
131072 & $ 1.46$ & $ 1.25$ \\
\hline
1B & $2 \times 4$ &
16384 & $ 1.46$ & $ 0.95$ \\
& &
32768 & $ 1.57$ & $ 1.07$ \\
& &
65536 & $ 1.65$ & $ 1.22$ \\
& &
98304 & $ 1.68$ & $ 1.30$ \\
& &
131072 & $ 1.70$ & $ 1.35$ \\
& &
196608 & $ 1.71$ & $ 1.40$ \\
& &
262144 & $ 1.72$ & $ 1.41$ \\
\hline
1B & $4 \times 2$ &
8192 & $ 1.22$ & $ 0.96$ \\
& &
16384 & $ 1.31$ & $ 1.06$ \\
& &
32768 & $ 1.38$ & $ 1.11$ \\
& &
49152 & $ 1.41$ & $ 1.15$ \\
& &
65536 & $ 1.43$ & $ 1.20$ \\
& &
98304 & $ 1.45$ & $ 1.23$ \\
& &
131072 & $ 1.46$ & $ 1.24$ \\
\hline

\end{tabular}

\begin{tabular}{rrr|r|r}
Model & Mesh & $n_\text{seq}$ & TMS & Real Speedup \\
\hline
3B & $2 \times 2$ &
8192 & $ 1.17$ & $ 0.96$ \\
& &
16384 & $ 1.26$ & $ 1.04$ \\
& &
32768 & $ 1.34$ & $ 1.11$ \\
& &
49152 & $ 1.38$ & $ 1.18$ \\
& &
65536 & $ 1.40$ & $ 1.22$ \\
& &
98304 & $ 1.43$ & $ 1.25$ \\
& &
131072 & $ 1.45$ & $ 1.27$ \\
\hline
3B & $2 \times 4$ &
16384 & $ 1.38$ & $ 0.94$ \\
& &
32768 & $ 1.51$ & $ 1.09$ \\
& &
65536 & $ 1.61$ & $ 1.23$ \\
& &
98304 & $ 1.65$ & $ 1.32$ \\
& &
131072 & $ 1.67$ & $ 1.36$ \\
& &
196608 & $ 1.69$ & $ 1.43$ \\
& &
262144 & $ 1.71$ & $ 1.45$ \\
\hline
3B & $4 \times 2$ &
8192 & $ 1.17$ & $ 0.97$ \\
& &
16384 & $ 1.26$ & $ 1.04$ \\
& &
32768 & $ 1.34$ & $ 1.13$ \\
& &
49152 & $ 1.38$ & $ 1.20$ \\
& &
65536 & $ 1.40$ & $ 1.22$ \\
& &
98304 & $ 1.43$ & $ 1.26$ \\
& &
131072 & $ 1.45$ & $ 1.27$ \\
\hline
\end{tabular}

\begin{tabular}{rrr|r|r}
Model & Mesh & $n_\text{seq}$ & TMS & Real Speedup \\
\hline
7B & $2 \times 2$ &
8192 & $ 1.15$ & $ 0.93$ \\
& &
16384 & $ 1.23$ & $ 1.03$ \\
& &
32768 & $ 1.31$ & $ 1.11$ \\
& &
49152 & $ 1.36$ & $ 1.16$ \\
& &
65536 & $ 1.38$ & $ 1.20$ \\
& &
98304 & $ 1.42$ & $ 1.23$ \\
& &
131072 & $ 1.43$ & $ 1.25$ \\
\hline
7B & $2 \times 4$ &
16384 & $ 1.34$ & $ 0.94$ \\
& &
32768 & $ 1.47$ & $ 1.07$ \\
& &
65536 & $ 1.58$ & $ 1.23$ \\
& &
98304 & $ 1.62$ & $ 1.30$ \\
& &
131072 & $ 1.65$ & $ 1.36$ \\
& &
196608 & $ 1.68$ & $ 1.39$ \\
& &
262144 & $ 1.70$ & $ 1.43$ \\
\hline
7B & $4 \times 2$ &
8192 & $ 1.15$ & $ 0.98$ \\
& &
16384 & $ 1.23$ & $ 1.03$ \\
& &
32768 & $ 1.31$ & $ 1.12$ \\
& &
49152 & $ 1.36$ & $ 1.18$ \\
& &
65536 & $ 1.38$ & $ 1.20$ \\
& &
98304 & $ 1.42$ & $ 1.23$ \\
& &
131072 & $ 1.43$ & $ 1.25$ \\
\end{tabular}

\subsection{TPU v3 8-chip results}
\begin{tabular}{rrr|r|r}
Model & Mesh & $n_\text{seq}$ & TMS & Speedup \\
\hline
1B & $2 \times 4$ &
16384 & $ 1.33$ & $ 1.12$ \\
& &
32768 & $ 1.46$ & $ 1.27$ \\
& &
65536 & $ 1.57$ & $ 1.41$ \\
& &
98304 & $ 1.62$ & $ 1.47$ \\
\hline
1B & $4 \times 2$ &
8192 & $ 1.14$ & $ 1.07$ \\
& &
16384 & $ 1.22$ & $ 1.16$ \\
& &
32768 & $ 1.31$ & $ 1.25$ \\
& &
49152 & $ 1.35$ & $ 1.30$ \\
\hline
3B & $2 \times 4$ &
16384 & $ 1.26$ & $ 1.12$ \\
& &
32768 & $ 1.38$ & $ 1.27$ \\
& &
65536 & $ 1.51$ & $ 1.41$ \\
& &
98304 & $ 1.57$ & $ 1.47$ \\
\hline
3B & $4 \times 2$ &
8192 & $ 1.10$ & $ 1.07$ \\
& &
16384 & $ 1.17$ & $ 1.16$ \\
& &
32768 & $ 1.26$ & $ 1.25$ \\
& &
49152 & $ 1.31$ & $ 1.30$ \\
\hline
\end{tabular}

\subsection{TPU v4 16-chip results}
\begin{tabular}{rrr|r|r}
Model & Mesh&  $n_\text{seq}$ & TMS & Speedup \\
\hline
1B & $2 \times 8$ &
32768 & $ 1.54$ & $ 1.06$ \\
& &
131072 & $ 1.76$ & $ 1.26$ \\
& &
262144 & $ 1.81$ & $ 1.37$ \\
& &
393216 & $ 1.83$ & $ 1.43$ \\
& &
524288 & $ 1.84$ & $ 1.47$ \\
& &
655360 & $ 1.85$ & $ 1.49$ \\
& &
786432 & $ 1.85$ & $ 1.51$ \\
\hline
1B & $4 \times 4$ &
16384 & $ 1.33$ & $ 1.06$ \\
& &
65536 & $ 1.57$ & $ 1.22$ \\
& &
131072 & $ 1.65$ & $ 1.31$ \\
& &
196608 & $ 1.68$ & $ 1.36$ \\
& &
262144 & $ 1.70$ & $ 1.39$ \\
& &
327680 & $ 1.71$ & $ 1.40$ \\
& &
393216 & $ 1.71$ & $ 1.42$ \\
\hline
\end{tabular}
\begin{tabular}{rrr|r|r}
Model & Mesh & $n_\text{seq}$ & TMS & Speedup \\
\hline
3B & $2 \times 8$ &
32768 & $ 1.45$ & $ 1.09$ \\
& &
131072 & $ 1.71$ & $ 1.36$ \\
& &
262144 & $ 1.78$ & $ 1.50$ \\
& &
393216 & $ 1.81$ & $ 1.57$ \\
& &
524288 & $ 1.83$ & $ 1.60$ \\
& &
655360 & $ 1.84$ & $ 1.63$ \\
& &
786432 & $ 1.84$ & $ 1.65$ \\
\hline
3B & $4 \times 4$ &
16384 & $ 1.26$ & $ 1.05$ \\
& &
65536 & $ 1.51$ & $ 1.22$ \\
& &
131072 & $ 1.61$ & $ 1.32$ \\
& &
196608 & $ 1.65$ & $ 1.38$ \\
& &
262144 & $ 1.67$ & $ 1.41$ \\
& &
327680 & $ 1.68$ & $ 1.43$ \\
& &
393216 & $ 1.69$ & $ 1.45$ \\
\hline
\end{tabular}
\begin{tabular}{rrr|r|r}
Model & Mesh & $n_\text{seq}$ & TMS & Speedup \\
\hline
7B & $2 \times 8$ &
32768 & $ 1.40$ & $ 1.07$ \\
& &
131072 & $ 1.67$ & $ 1.30$ \\
& &
262144 & $ 1.76$ & $ 1.44$ \\
& &
393216 & $ 1.80$ & $ 1.51$ \\
& &
524288 & $ 1.81$ & $ 1.56$ \\
& &
655360 & $ 1.83$ & $ 1.59$ \\
\hline
7B & $4 \times 4$ &
16384 & $ 1.22$ & $ 1.04$ \\
& &
65536 & $ 1.47$ & $ 1.18$ \\
& &
131072 & $ 1.58$ & $ 1.28$ \\
& &
196608 & $ 1.62$ & $ 1.33$ \\
& &
262144 & $ 1.65$ & $ 1.37$ \\
& &
327680 & $ 1.67$ & $ 1.40$ \\
& &
393216 & $ 1.68$ & $ 1.41$ \\
\end{tabular}


\end{document}